%% file: main.tex
\pdfoutput=1
\documentclass[11pt]{article}

\usepackage[]{acl}

\usepackage{times}
\usepackage{latexsym}
\usepackage[T1]{fontenc}
\usepackage[utf8]{inputenc}
\usepackage{microtype}

\usepackage{xspace}
\usepackage{graphicx}
\usepackage{multirow}
\usepackage{tcolorbox}
\usepackage{booktabs}
\usepackage{amsmath}
\usepackage{amssymb}
\usepackage[normalem]{ulem}
\usepackage{xcolor}
\usepackage{colortbl}
\usepackage{utfsym}
\usepackage{pifont}
\usepackage{wasysym}
\usepackage{makecell}
\usepackage{fontawesome}
\usepackage{threeparttable}
\usepackage{enumitem}

\usepackage{longtable}

\def\eg{\textit{e.g.,} }
\def\ie{\textit{i.e.,} }
\def\bench{DevEval\xspace}

\title{\bench: A Manually-Annotated Code Generation Benchmark Aligned with Real-World Code Repositories}

\author{
    \makecell{Jia Li $\male^{1,2}$, Ge Li$^{1,2}$, Yunfei Zhao$^{1,2}$, Yongmin Li$^{1,2}$, Huanyu Liu$^{1,2}$, Hao Zhu$^{1,2}$, Lecheng Wang$^{1,2}$ \\ 
    Kaibo Liu$^{1,2}$, Zheng Fang$^{1,2}$, Lanshen Wang$^{1,2}$, Jiazheng Ding$^{1,2}$, Xuanming Zhang$^{1,2}$ \\
    Yuqi Zhu$^{1,2}$, Yihong Dong$^{1,2}$, Zhi Jin$^{1,2}$, Binhua Li$^3$, Fei Huang$^3$, Yongbin Li$^3$ \\} \\
    $^1$School of Computer Science, Peking University \\
    $^2$ Key Laboratory of High Confidence Software Technologies (Peking University), Ministry of Education \\
    $^3$Alibaba Group  \\ 
    \texttt{lijia@stu.pku.edu.cn, \{lige, zhijin\}@pku.edu.cn} \\
}

\begin{document}

\maketitle

\begin{abstract}
How to evaluate the coding abilities of Large Language Models (LLMs) remains an open question. We find that existing benchmarks are poorly aligned with real-world code repositories and are insufficient to evaluate the coding abilities of LLMs.

To address the knowledge gap, we propose a new benchmark named \textbf{\bench}, which has three advances. 
\ding{182} \bench aligns with real-world repositories in multiple dimensions, \eg code distributions and dependency distributions. \ding{183} \bench is annotated by 13 developers and contains comprehensive annotations (\eg requirements, original repositories, reference code, and reference dependencies). 
\ding{184} \bench comprises 1,874 testing samples from 117 repositories, covering 10 popular domains (\eg Internet, Database).
Based on \bench, we propose \textbf{repository-level code generation} and evaluate 8 popular LLMs on \bench (\eg gpt-4, gpt-3.5, StarCoder 2, DeepSeek Coder, CodeLLaMa).
Our experiments reveal these LLMs' coding abilities in real-world code repositories. 
\textbf{For example, the highest Pass@1 of gpt-4-turbo only is 53.04\% in our experiments.} 
We also analyze LLMs' failed cases and summarize their shortcomings. We hope \bench can facilitate the development of LLMs in real code repositories. \bench, prompts, and LLMs' predictions have been released\footnote{\url{https://github.com/seketeam/DevEval}}.
\end{abstract}

\input{chapter/Introduction}

\input{chapter/Benchmark}

\input{chapter/Benchmark_Collection}

\input{chapter/Experiments}

\input{chapter/Discussion}

\input{chapter/Related_Work}

\input{chapter/Conclusion}

\bibliography{anthology,custom}
\bibliographystyle{acl_natbib}

% \appendix

% \input{chapter/Appendix}

\end{document}

%% file: chapter/Introduction.tex
\section{Introduction}
\label{sec:Introduction}

Code generation with Large Language Models (LLMs) has attracted lots of researchers' attention \cite{DeepSeek_Coder,CodeLLaMa,StarCoder-2}, and some commercial products have been produced, \eg GitHub Copilot \cite{Copilot}. 
With more and more LLMs emerging, how to evaluate LLMs on code generation remains an open question. 

\begin{figure}[t]
\centering
\includegraphics[width=0.9\linewidth]{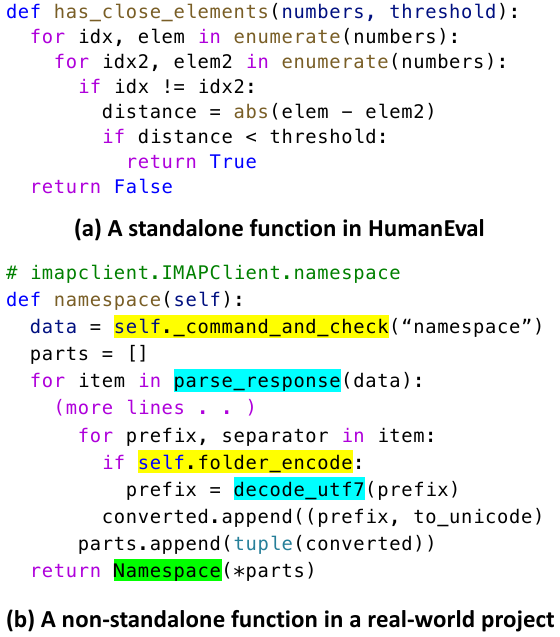}
\caption{Examples of standalone and non-standalone functions. Dependencies are highlighted, \ie yellow: intra-class dependencies, green: intra-file dependencies, and blue: cross-file dependencies.}
\label{fig:Introduction_Example}
\end{figure}

Existing benchmarks are mainly composed of hand-crafted programming problems and are poorly aligned with real-world code repositories. 
LLMs' performance on these benchmarks is inconsistent with developers' actual experiences in real-world software development.
Thus, a benchmark aligned with real-world repositories is necessary. We analyze over 1 million functions from 500 real-world repositories (see Section \ref{sec:benchmark_collection}) and think a good benchmark should satisfy the following features.

\input{table/feature_comparison}

\begin{itemize}[leftmargin=*]
    \item \textbf{Real-world Repository.} 
    The benchmark should be collected from real-world code repositories \cite{CoderEval}.
    \item \textbf{Real Code Distribution.} Real-world repositories comprise two types of code, \ie standalone and non-standalone code.
    As shown in Figure \ref{fig:Introduction_Example}, a standalone function solely uses built-in or public libraries, while a non-standalone one contains context-aware \textit{dependencies} (\ie invocations of code elements defined in current repositories). The benchmark should cover both types of code and ensure their ratios are realistic. The number of dependencies should also be consistent with real-world repositories.
    \item \textbf{Comprehensive Annotations.} The benchmark can offer comprehensive annotations, including natural language requirements, original repositories, and ground truths (code and dependencies).
    \item \textbf{Robust Evaluation Metrics.} The benchmark should provide execution-based metrics (\eg Pass@$k$) evaluate functional correctness of programs and metrics to assess the accuracy of dependencies in programs.
\end{itemize}
However, as shown in Table \ref{tab:feature_comparison}, none of the existing benchmarks satisfies all aforementioned features.
The problem hinders the evaluation and development of LLMs in the real development process.

\textbf{To address the above problem, we propose a new code generation benchmark named \bench, which aligns with real-world code repositories.} 
As shown in Table \ref{tab:feature_comparison}, \bench satisfies the above features. 
\ding{182} \bench comprises 1,874 testing samples from 117 real-world repositories, which cover 10 popular domains (\eg Internet, Database).
\ding{183} \bench is constructed through a rigorous pipeline and aligns with real-world repositories. Specifically, the distributions of code and dependencies in \bench are consistent with the ones in 500 real-world repositories. Detailed statistics are in Section \ref{sec:DevEval:features}. 
\ding{184} \bench is annotated by 13 developers and contains comprehensive annotations, \eg detailed requirements, original repositories, reference code, and reference dependencies. 
\ding{185} \bench leverages test cases to check models' predictions and report Pass@$k$. It also proposes Recall@$k$ to evaluate the dependencies in predictions.

Based on \bench, we propose \textbf{repository-level code generation}, which simulates the developers' coding process in a working repository. The task asks models to write the code based on requirements and a complete repository.

We evaluate 8 popular LLMs (\ie gpt-4 \cite{gpt-4}, gpt-3.5 \cite{gpt-3.5}, DeepSeek Coder \cite{DeepSeek_Coder}, StarCoder 2 \cite{StarCoder-2}, CodeLLaMa \cite{CodeLLaMa}). These LLMs exhibit low performance on \bench, especially compared to their performance on previous benchmarks. \textbf{For example, gpt-4-turbo-1106 achieves a Pass@1 score of 80\% on HumanEval, while its highest Pass@1 on \bench is only 53.04\%.} Our results reveal the coding abilities of these LLMs in real-world repositories. We further analyze failed cases and summarize the shortcomings of existing LLMs in \bench.

\begin{figure*}[t]
\centering
\includegraphics[width=0.85\linewidth]{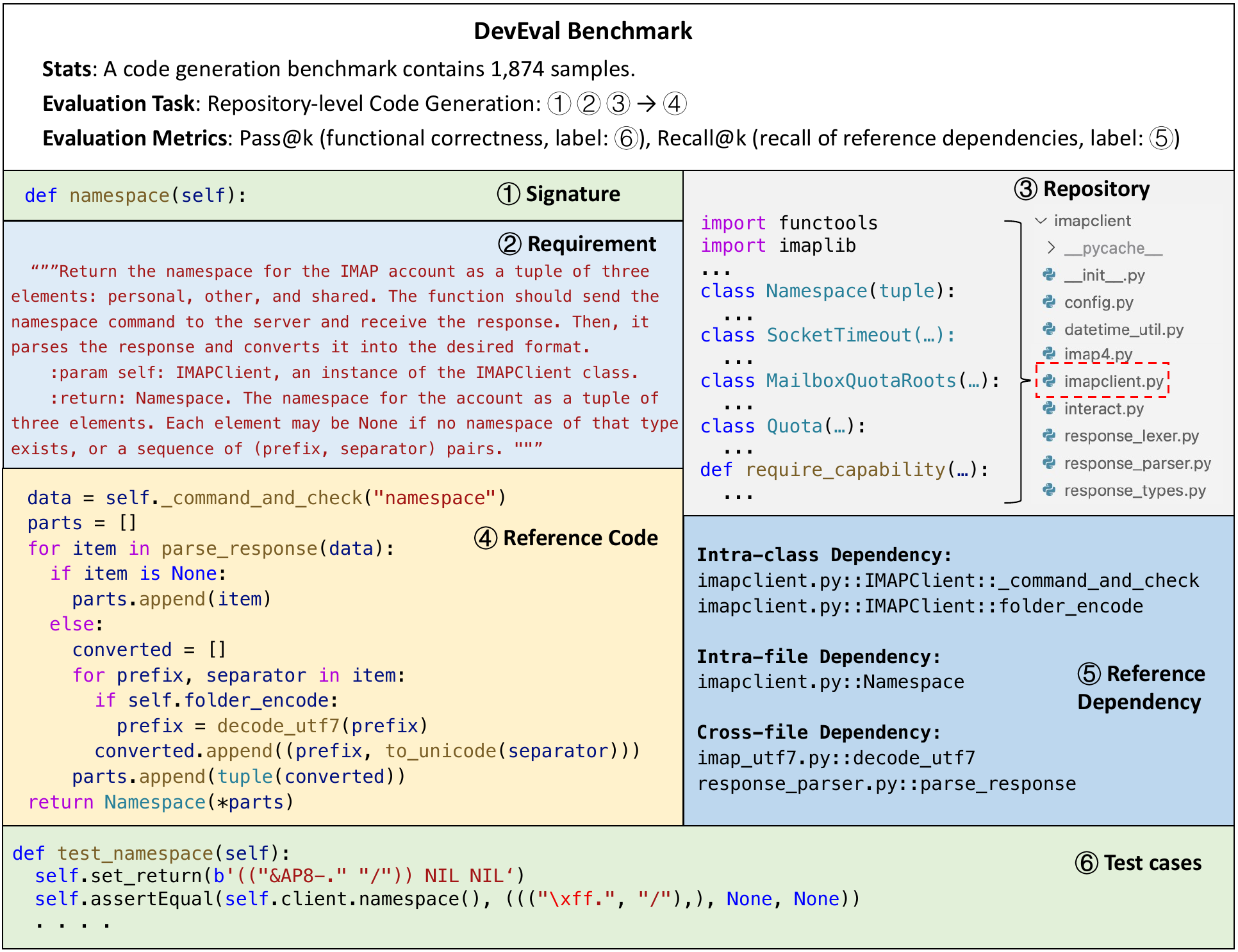}
\caption{An overview of \bench. Each sample consists of six components.}
\label{fig:Benchmark_example}
\end{figure*}

In summary, our contributions are as follows:
\begin{itemize}[leftmargin=*]
    \item We summarize four features (see Table \ref{tab:feature_comparison}) that a code generation benchmark for real-world repositories should satisfy.
    \item We propose a new code generation benchmark - \bench, satisfying the above features. The benchmark has been released.
    \item We propose repository-level code generation, which provides a challenging and realistic evaluation scenario.
    \item We evaluate 8 popular LLMs on \bench, analyzing their strengths and shortcomings in repository-level code generation.
\end{itemize}

We hope \bench can align with the actual experiences of developers during the practical development process. By \bench, practitioners can pick up superior LLMs and facilitate the application of code generation techniques in real-world repositories.

%% file: table/feature_comparison.tex
\begin{table*}[t]
\centering
\caption{The comparison between existing benchmarks and \bench.}
\label{tab:feature_comparison}
\resizebox{0.95\linewidth}{!}{
\begin{tabular}{lcccc}
\toprule
Benchmark & Real Repo. & Real Code Distribution & Comprehensive Annota. & Robust Metric \\ \midrule
CoNaLA \cite{CoNaLA} & \faTimes & \faTimes & \faTimes & \faTimes \\
Concode \cite{Concode} & \faCheck & \faTimes & \faTimes & \faTimes \\
HumanEval \cite{Codex} & \faTimes & \faTimes & \faTimes & \faTimes  \\
MBPP \cite{MBPP} & \faTimes & \faTimes & \faTimes & \faTimes  \\
APPS \cite{APPS} & \faTimes & \faTimes & \faTimes & \faTimes  \\
PandasEval \cite{CERT} & \faTimes & \faTimes & \faTimes & \faTimes  \\
NumpyEval \cite{CERT} & \faTimes & \faTimes & \faTimes & \faTimes  \\
AixBench \cite{SkCoder} & \faCheck & \faTimes & \faTimes & \faTimes  \\
ClassEval \cite{ClassEval} & \faTimes & \faTimes & \faTimes & \faTimes  \\
CoderEval \cite{CoderEval} & \faCheck & \faTimes & \faTimes & \faCheck  \\ \midrule
\bench (Ours) & \faCheck & \faCheck & \faCheck & \faCheck \\
\bottomrule
\end{tabular}}
\end{table*}

%% file: chapter/Benchmark.tex
\section{Benchmark - \bench}
\label{sec:DevEval}

\input{table/Benchmark_Comparison}

\input{table/Dependency_Distribution}

\subsection{Overview}
\label{sec:DevEval:overview}

\bench contains 1,874 samples derived from 117 real-world code repositories. 
As shown in Figure \ref{fig:Benchmark_example}, each sample consists of six components. 
\textbf{\ding{182} Function Signature:} The signature of the target code. 
\textbf{\ding{183} Requirement:} An English description detailing the functionality of the target code. 
\textbf{\ding{184} Repository Contexts:} Code contexts (\eg classes, functions, variables) defined outside the target code in the current repository.
\textbf{\ding{185} Reference Code:} A developer-written implementation of the target code. This code may invoke dependencies defined in the current repository.
\textbf{\ding{186} Reference Dependency:} The dependencies invoked in the reference code include intra-class, intra-file, and cross-file dependencies.
\textbf{\ding{187} Test Cases:} Test cases are used to check the functional correctness of the code.

\subsection{Task Definition}
\label{sec:DevEval:tasks}

Based on \bench, we propose \textbf{repository-level code generation} task. A model is given a function signature, a requirement, and a complete repository. The model is asked to output a function to satisfy the requirement. We then insert the function into its repository and check its correctness.

\subsection{Evaluation Metrics}
\label{sec:DevEval:metric}

\textbf{Pass@$k$ (Functional Correctness).} Following previous studies \cite{Codex,MBPP,CoderEval}, we assess the functional correctness of programs by executing test cases and compute the unbiased Pass@$k$.
Specifically, we generate $n \geq k$ programs per requirement, count the number of correct programs $c \leq n$ that pass test cases, and calculate the Pass@$k$:
\begin{equation}
\text{Pass}@k:=\underset{\text { Requirements }}{\mathbb{E}}\left[1-\frac{\left(\begin{array}{c}
n-c \\
k
\end{array}\right)}{\left(\begin{array}{l}
n \\
k
\end{array}\right)}\right]
\end{equation}

\textbf{Recall@$k$ (Recall of Reference Dependency).} 
Besides the functional correctness, we expect LLMs to invoke relevant dependencies defined in contexts. Hence, we propose Recall@$k$, which gauges the recall of reference dependencies in generated programs.  

Specifically, LLMs generate $k$ programs per requirement. For the $i$-th program, we employ a parser\footnote{We develop the parser based on an open-source static analysis tool - Pyan \cite{Pyan}.} to extract its dependencies as $\mathbb{P}_i$. Subsequently, we compare $\mathbb{P}_i$ with reference dependencies $\mathbb{R}$ and compute the Recall@$k$:
\begin{equation}
\text{Recall}@k:= \underset{\text{Requirements}}{\mathbb{E}} \left[ \max_{i \in [1, k]} \frac{|\mathbb{R} \cap \mathbb{P}_i|}{|\mathbb{R}|}\right]
\end{equation}
where $|\cdot|$ means the number of elements of a set.

\subsection{Features of \bench}
\label{sec:DevEval:features}

Compared to existing benchmarks, \bench shows three unique advances, which we discuss below.

\ding{182} \textbf{Alignment with real-world code repositories.} Table \ref{tab:benchmark_comparison} shows the data distributions of existing benchmarks and \bench. We also show the data distributions of 500 real-world repositories and consider them as the oracle. We can see that \bench aligns with 500 real repositories in multiple aspects, \ie code distributions, the number of dependencies, and the scale of repositories. Table \ref{tab:dependency_distribution} further shows the distribution of dependency types, \ie intra-class, intra-file, and cross-file dependencies. 
\bench outperforms previous benchmarks in all types, showing a distribution that is close to the distribution in 500 real repositories.

\ding{183} \textbf{Comprehensive Annotations.} As shown in Figure \ref{fig:Benchmark_example}, \bench provides comprehensive annotations that are labeled by 13 human developers. Particularly, \bench has advantages in requirements and reference dependencies. 

\textit{Requirements.} Original code comments in repositories often are vague and are different from requirements in practice. We engaged 13 developers to write requirements, costing approximately 674 person-hours manually. As depicted in Figure \ref{fig:Benchmark_example}, each requirement encapsulates the code's functionality and input-output parameters. The average length of requirements in \bench (91.5 tokens) more than doubles that of CoderEval (41.5 tokens). 
\textit{Reference Dependencies.} Previous benchmarks (\ie CoderEval, ClassEval) only provide dependencies' names (\eg \texttt{close}). Because many functions have the same name in practice, it is hard to identify whether generated dependencies are correct by relying on names. \bench annotates dependencies with paths (\eg \texttt{A.py::ClassB::close}), addressing ambiguity and biases. These annotations offer a broad arena to explore repository-level code generation and evaluation.

\ding{184} \textbf{A realistic task and evaluation metrics.} Traditional benchmarks fall into a simple requirement-to-code task. In contrast, \bench proposes a more realistic task - repository-level code generation. This task simulates the coding process of developers in a working repository. Besides, we design two metrics to comprehensively assess the correctness of generated programs in functionality and dependencies.

\ding{185} \textbf{Wide scope for research communities.} The \bench and repository-level code generation can serve as an arena to compare approaches ranging from retrieval and long-context models to decision-making agents. \bench also allows creative freedom, as models can generate diverse programs to meet requirements.

%% file: table/Benchmark_Comparison.tex
\begin{table*}[t]
\caption{The comparison between popular code generation benchmarks and \bench. SA: Standalone. L(Re): the average lengths (tokens) of requirements.}
\label{tab:benchmark_comparison}
\vspace{-0.2cm}
\resizebox{\linewidth}{!}{
\begin{tabular}{l|cccc|cccc|cc|c}
\toprule
\multirow{2}{*}{Benchmark} & \multicolumn{4}{c|}{Code Distribution} & \multicolumn{4}{c}{Dependency} & \multicolumn{2}{|c|}{Repository's Scale} & \multirow{2}{*}{\#L(Re)} \\
 & \#Repo & \#Total & SA (\%) & Non-SA (\%) & \#Type & \#Total & \#Per Sample & Path & \#File & \#Line & \\ \midrule
CoNaLA \cite{CoNaLA} & -- & 500 & 100\% & 0\% & 0 & 0 & 0 & \usym{2717} & 0 & 0 & 13.1 \\
HumanEval \cite{Codex} & -- & 164 & 100\% & 0\% & 0 & 0 & 0 & \usym{2717} & 0 & 0 & 58.8 \\
MBPP \cite{MBPP} & -- & 974 & 100\% & 0\% & 0 & 0 & 0 & \usym{2717} & 0 & 0 & 16.1 \\
PandasEval \cite{CERT} & -- & 101 & 100\% & 0\% & 0 & 0 & 0 & \usym{2717} & 0 & 0 & 29.7 \\
NumpyEval \cite{CERT} & -- & 101 & 100\% & 0\% & 0 & 0 & 0 & \usym{2717} & 0 & 0 & 30.5 \\
AixBench \cite{SkCoder} & -- & 175 & 100\% & 0\% & 0 & 0 & 0 & \usym{2717} & 0 & 0 & 34.5 \\
ClassEval \cite{ClassEval} & -- & 100 & 100\% & 0\% & 0 & 0 & 0 & \usym{2717} & 0 & 0 & -- \\
\midrule
Concode \cite{Concode} & -- & 2,000 & 20\% & 80\% & 1 & 2,455 & 1.23 & \usym{2717} & 0 & 0 & 16.8 \\
CoderEval \cite{CoderEval} & 43 & 230 & 36\% & 64\% & 3 & 256 & 1.73 & \usym{2717} & 71 & 14,572 & 41.5 \\
\rowcolor[rgb]{ .741,  .843,  .933}
\textbf{\bench} & \textbf{117} & \textbf{1,874} & \textbf{27\%} & \textbf{73\%} & \textbf{3} & \textbf{4,672} & \textbf{3.41} & \usym{2713} & \textbf{243} & \textbf{45,941} & \textbf{91.5} \\ \midrule
500 Real-world Repos & 500 & 1M & 27\% & 73\% & 3 & 3M & 3.22 & -- & 238 & 46,313 & -- \\ \bottomrule
\end{tabular}}
\end{table*}

%% file: table/Dependency_Distribution.tex
\begin{table}[t]
\caption{The distribution of dependency types. The values in parentheses are the corresponding percentages in all dependencies.}
\label{tab:dependency_distribution}
\vspace{-0.2cm}
\resizebox{\linewidth}{!}{
\begin{tabular}{lccccc}
\toprule
\multicolumn{1}{c}{\begin{tabular}[c]{@{}c@{}}Dependency \\ Type\end{tabular}} & HumanEval & Concode & CoderEval & \cellcolor[rgb]{ .741,  .843,  .933}\bench & 500 Projects \\ \midrule
Intra-class & 0 & 2,455 (100\%) & 117 (46\%) & \cellcolor[rgb]{ .741,  .843,  .933}\textbf{1,778 (38\%)} & 939k (42\%) \\
Intra-file & 0 & 0 & 90 (35\%) & \cellcolor[rgb]{ .741,  .843,  .933}\textbf{1,502 (32\%)} & 597k (29\%) \\
Cross-file & 0 & 0 & 49 (19\%) & \cellcolor[rgb]{ .741,  .843,  .933}\textbf{1,392 (30\%)} & 611k (30\%) \\
\bottomrule
\end{tabular}}
\end{table}

%% file: chapter/Benchmark_Collection.tex
\section{Benchmark Construction}
\label{sec:benchmark_collection}

\begin{figure}[t]
\centering
\includegraphics[width=0.9\linewidth]{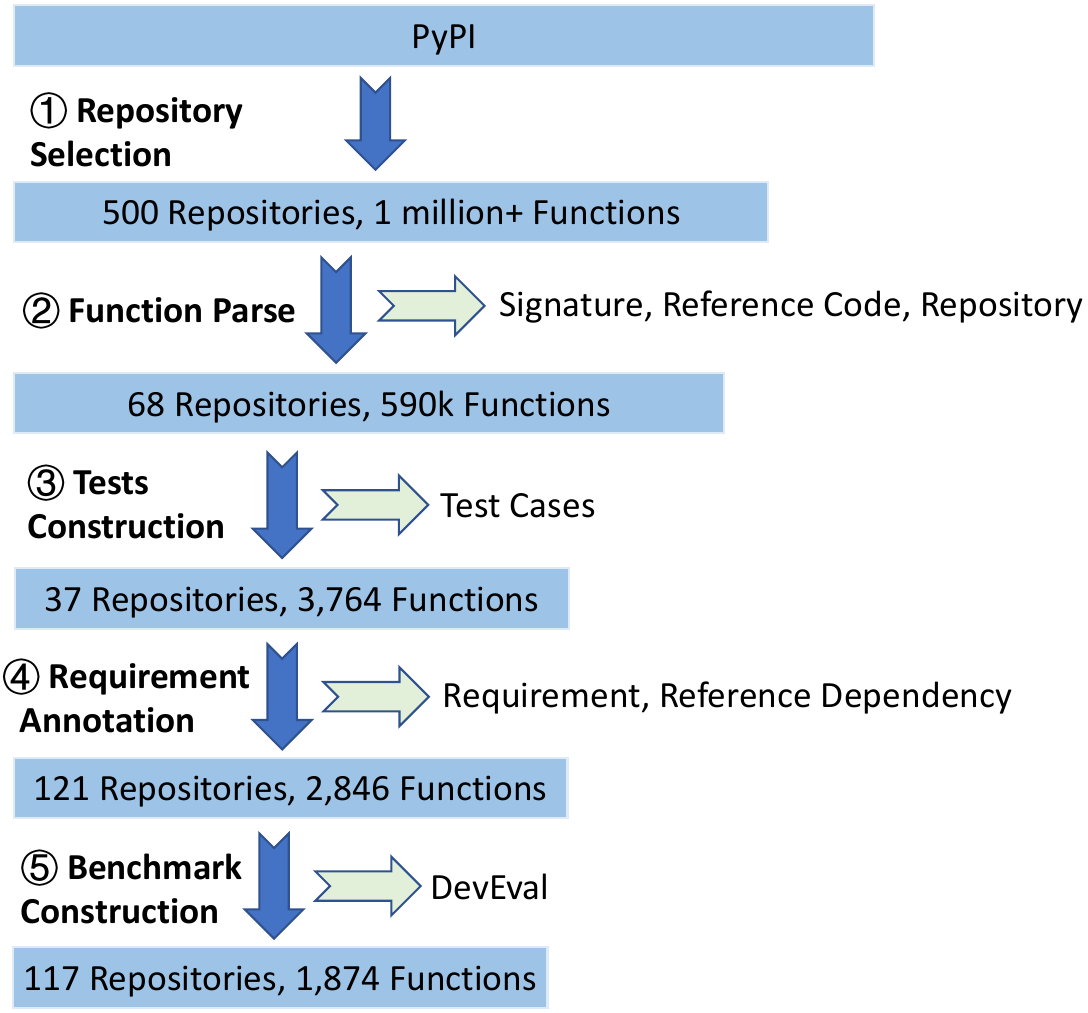}
\caption{The process of collecting \bench.}
\label{fig:Benchmark_contruction}
\end{figure}

As shown in Figure \ref{fig:Benchmark_contruction}, the collection of \bench consists of five stages.

\textbf{Stage \ding{182}: Repository Selection.} PyPI \cite{PyPI} is a rich data source for real code repositories. We identify the top 10 popular programming topics in PyPI and select the top 50 repositories under each topic. The selection follows three criteria: open-source licenses, non-fork and non-malicious repositories, and explicit unit
tests. We download the latest released versions in November 2023 and finally obtain 500 practical projects (10 topics * 50 projects).

\textbf{Stage \ding{183}: Function Parse.} We extract functions from 500 repositories and exclude trivial functions (\ie empty or initialization functions).
We extract each function's signature and function body (\ie reference code). The other programs with current repositories are considered as \textit{repository contexts}. Finally, this stage obtain 590,365 candidate functions.

\textbf{Stage \ding{184}: Tests Construction.} We extract test cases from repositories invoking specific candidate functions. We use a public framework - \texttt{setuptools} to automatically install running environments and run test cases with a popular testing framework - \texttt{Pytest}. Candidate functions without executable test cases are excluded. Meanwhile, we ensure our test cases succeed in the reference code and fail in the wrong programs. Finally, we retain 3,764 candidate functions.

\textbf{Stage \ding{185}: Human Annotation.} 
We engaged 13 developers to manually annotate requirements and reference dependencies for each candidate function. Given their countries of residence, all annotators obtain adequate payments.

Through discussions with annotators, we establish two criteria for requirements. \textit{Naturalness}--ensuring the requirement reads like a natural description from the perspective of a real-world developer. \textit{Functionality}--demanding clear descriptions of the code's purposes and input-output parameters. Each requirement undergoes a dual-annotation process, with one annotator assigned to its initial drafting and another responsible for a meticulous double-check. Trivial functions (\eg shortcut functions) and functions violating the ethical code (\eg malware) are excluded.
Subsequently, the same 13 annotators review the reference code and label its reference dependencies. Finally, we retain 2,846 functions with high-quality requirements and reference dependencies.

\textbf{Stage \ding{186}: Benchmark Construction.} We select candidate functions to construct based on two criteria: consistent with the data distribution of 500 real-world repositories and including as many functions as possible. Finally, we select 1,874 (73\%) non-standalone functions and 706 (27\%) standalone functions to construct \bench.

%% file: chapter/Experiments.tex
\section{Experiments}
\label{sec:experiments}

\input{table/Base_LLMs}

\subsection{Studied LLMs}
\label{sec:experiments:base_llms}

As shown in Table \ref{tab:Base_LLMs}, we evaluate 8 popular LLMs, including two closed-source models and six open-source models. We use official interfaces or implementations to reproduce these LLMs.

\input{table/Main_Result}

\subsection{Experimental Settings}
\label{sec:experiments:setting}

Repository-level code generation takes a requirement and a repository as inputs. Typically, a repository consists of hundreds of code files and is very long. For example, the average length of 500 real-world repositories is 1.1 million tokens, surpassing the context windows of existing LLMs (\eg gpt-4: 128k tokens). Inspired by related work \cite{Repo_prompt}, we try to extract parts of code contexts from the repository as inputs and design the following experimental settings.

\textbf{\ding{182} Without context.} In this setting, we ignore contexts and directly generate the code based on requirements and signatures.

\textbf{\ding{183} Local File (Completion).} The local file denotes the code file where the reference code is in. This setting simulates the scenario where developers continue to write code at the end of a file. Thus, we consider code snippets above the reference code in the local file as contexts. Then, LLMs generate code in an autoregressive manner based on requirements, signatures, and contexts.

\textbf{\ding{184} Local File (Infilling).} Different from the Local File (Completion) setting, this setting simulates the scenario where developers infill code in the middle of a file. Thus, we use the code snippets above and below the reference code in the local file as contexts. We evaluate LLMs that support code infilling and construct input sequences using official formats.

\subsection{Evaluation}
\label{sec:experiment:evaluation}

We use Pass@$k$ and Recall@$k$ (see Section \ref{sec:DevEval:metric}) to assess generated programs. In this paper, $k \in [1, 3, 5, 10]$. When $k=1$, we use the greedy search and generate a single program per requirement. When $k>1$, we use the nucleus sampling with a temperature 0.4 and sample 20 programs per requirement. We set the top-$p$ to 0.95 and the max generation length to 500.

\subsection{Main Results}
\label{sec:experiments:main_results}

The Pass@$k$ and Recall@$k$ of different LLMs in three experimental settings are shown in Table \ref{tab:main_results}.

\noindent \textbf{Without Context.} gpt-4 and DeepSeek Coder achieve the highest Pass@1 and Recall@1 among all LLMs, respectively. However, all LLMs exhibit relatively low Pass@$k$ and Recall@$k$ values compared to their performance on previous benchmarks. For instance, gpt-4 achieves a Pass@1 score of 88.4 on HumanEval, whereas it scores 17.40 on Pass@1 in this setting. The decreases validate our motivation that existing benchmarks can not comprehensively assess the coding abilities of LLMs in real-world repositories. Furthermore, the results emphasize the importance of contexts.

\noindent \textbf{Local File (Completion) and (Infilling).} After introducing the contexts within local files, the Pass@$k$ and Recall@$k$ of all LLMs obviously increase. For example, the Pass@1 of gpt-4 is improved by 205\% and 173\% in two settings, respectively. 

\textbf{Successful Case Analyses.} We further inspect successful cases of gpt-4 and attribute the improvements to the synergy of contexts and requirements.
On the one hand, the contexts provide lots of domain knowledge. For example, the local file contains essential local environments (\eg current classes, imported libraries) and a majority of dependencies (\eg intra-class and intra-file: 72\% in \bench). Recent work \cite{RepoCoder,CrossCodeEval} in code completion also proved the importance of contexts. On the other hand, our manually written requirements elaborate on the code's purposes and the repositories' background knowledge. Thus, the requirements help LLMs understand long contexts and locate relevant dependencies.

\begin{figure}[t]
\centering
\includegraphics[width=\linewidth]{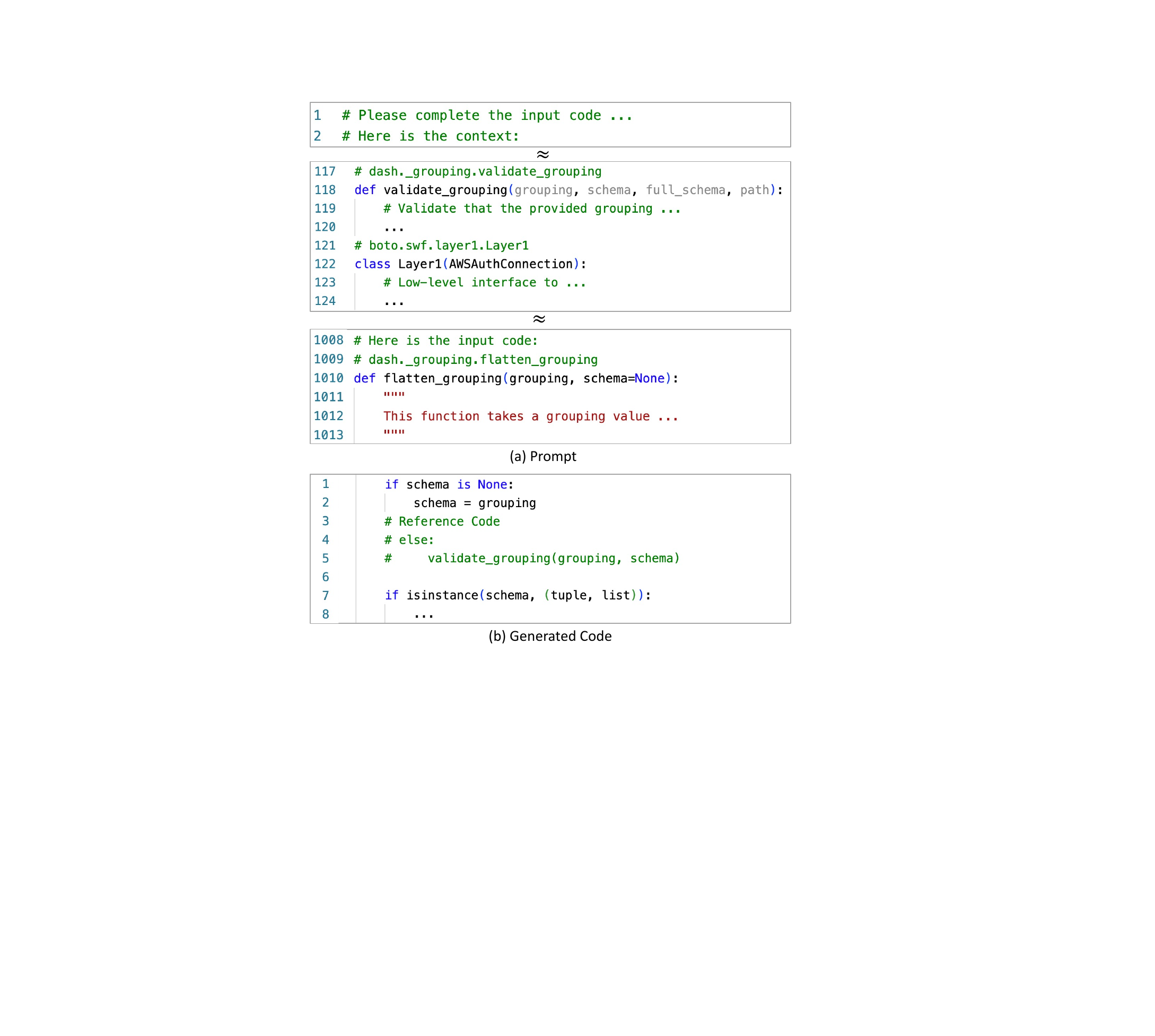}
\caption{A failed case of gpt-3.5 in the Local File (Completion) setting.}
\label{fig:error_case}
\end{figure}

\textbf{Error Case Analyses.}
Although promising, LLMs' performance in repository-level code generation is not satisfying. A manual inspection of failed cases reveals LLMs struggle with understanding contexts. Figure \ref{fig:error_case} illustrates a failed case. LLMs invoke a non-existent function - \texttt{create\_connection}, even though a valid function \texttt{connect} is present in the contexts. 
We think two reasons cause this problem. 

\textit{First, the contexts are too long.} The complete repositories are lengthy, approximately 9 times the context window of the state-of-the-art LLM - gpt-4-1106. Even when partial contexts are considered, their lengths match or exceed most current LLMs' context windows. Recent work \cite{Lost_in_the_Middle} has found that LLMs often ignore relevant information in the middle of long contexts. This finding is consistent with our results.
\textit{Second, the contexts are heterogeneous.} In other words, the contexts are composed of discrete code snippets from different files rather than a continuous file. As shown in Figure \ref{fig:error_case}, the programs within contexts come from multiple files, \eg \texttt{boto.regioninfo.py} and \texttt{boto.swf.layer1.py}. However, LLMs are typically trained to predict the next tokens based on the continuous contexts. The gap between training and inference objectives leads to a poor understanding of LLMs in contexts. Recent work \cite{ICL_Pretrain} also found similar gaps in reading comprehension and question answering.

\begin{figure}[t]
\centering
\includegraphics[width=\linewidth]{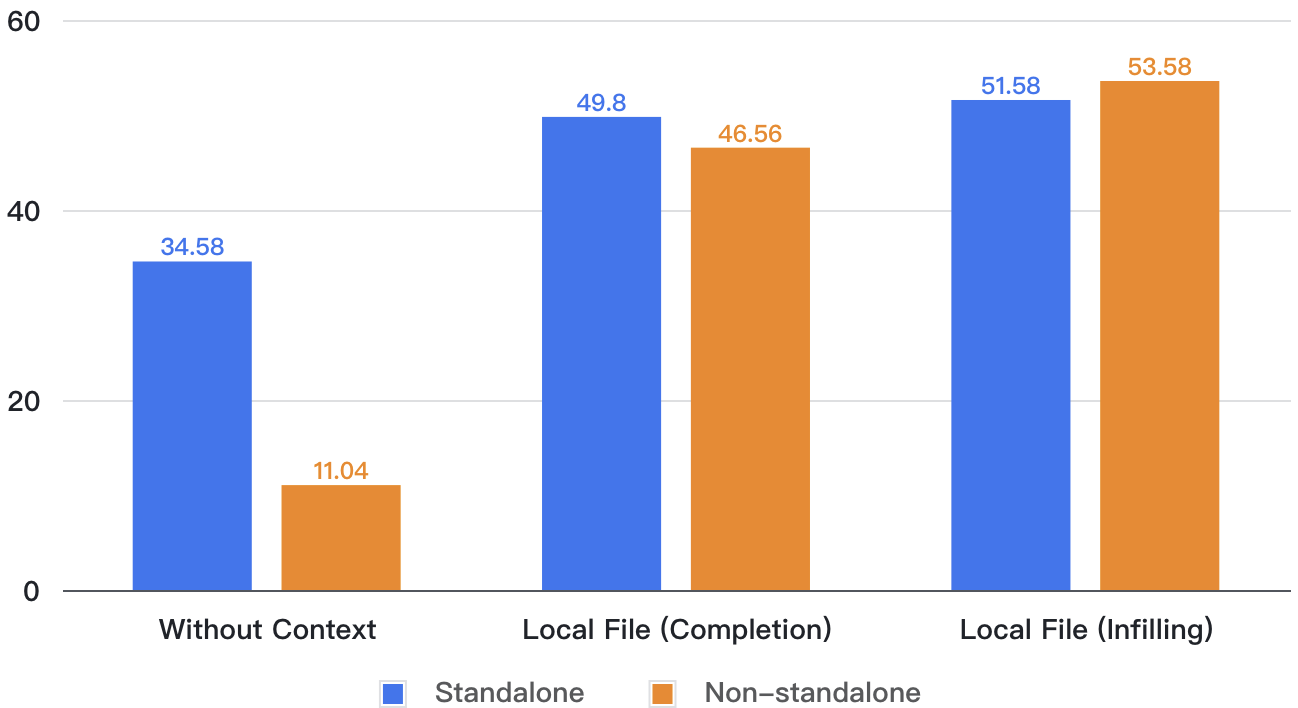}
\caption{Pass@1 of gpt-4 on different program types.}
\label{fig:different_program}
\end{figure}

We also obtain some interesting findings from Table \ref{tab:main_results}.

\ding{182} \textbf{LLMs successfully generate some dependencies without context.} Theoretically, LLMs do not see the contexts and cannot generate dependencies defined in the contexts. According to Table \ref{tab:main_results}, we are surprised to find that LLMs are able to generate some dependencies without context. We manually inspect successful cases and summarize two reasons. First, LLMs can reason about some easy dependencies from requirements, \eg initialization functions of returned objects. Second, LLMs can ``guess`` dependencies from their functionalities. In practice, dependencies' names come from their functional descriptions, \eg \texttt{send\_request()} - send a request to the server. LLMs are trained with a large code corpus and can learn the naming conventions. Thus, LLMs may successfully guess some dependencies from their functionalities.

\ding{183} \textbf{More contexts benefit code generation.}
Based on Table \ref{tab:main_results}, we compare the performance of an LLM (\eg gpt-4) under different settings. Obviously, the more input contexts, the better the performance of the LLM. It inspires practitioners to extend the context windows of LLMs and input more contexts.

\ding{184} \textbf{In the without context setting, gpt family models have higher Pass@$k$ and lower Recall@$k$, while other models are the opposite.} We speculate the reason is that gpt family models are instruction-tuned models and focus on performing tasks based on given instructions. With limited contexts, gpt family models are conservative and tend to generate code independently. Other LLMs are fundamental language models trained with real code files containing dependencies. They are aggressive and generate dependencies that may exist. The comparisons show the importance of instruction tuning in practical applications.

\subsection{Empirical Lessons}
\label{sec:experiment:lesson}

Based on the above experiments, we summarize the empirical lessons we learned as

\ding{182} \bench poses new challenges, \ie repository-level code generation. The performance of existing LLMs on \bench drops dramatically compared to their performance on previous benchmarks.

\ding{183} LLMs benefit from code contexts in current repositories. With limited context windows, the contexts from local files can improve gpt-4 by 205\% in Pass@1.

\ding{184} Detailed and accurate requirements help LLMs know the purposes of programs and understand long contexts.

\ding{185} LLMs struggle with understanding long and heterogeneous contexts. It causes LLMs to disregard the knowledge in contexts and even generate hallucinations (\eg non-existent functions).

%% file: table/Base_LLMs.tex
\begin{table}[t]
\caption{Studied LLMs in this paper. Context L.: Context Window.}
\label{tab:Base_LLMs}
\resizebox{\linewidth}{!}{
\begin{tabular}{llcc}
\toprule
Type & Name & Version & Context W. \\ \midrule
\multirow{2}{*}{Closed-source} 
 & gpt-4 & gpt-4-turbo-1106 & 128,000 \\ 
 & gpt-3.5 & gpt-3.5-turbo-1106 & 16,385 \\ \midrule
\multirow{6}{*}{Open-source} 
 & StarCoder 2 & 15B & 16,384 \\
 & StarCoder 2 & 7B & 16,384 \\
 & DeepSeek Coder & 33B & 16,384 \\
 & DeepSeek Coder & 6.7B & 16,384 \\
 & CodeLLaMa & 13B & 16,384 \\
 & CodeLLaMa & 7B & 16,384 \\
 \bottomrule
\end{tabular}}
\end{table}

%% file: table/Main_Result.tex
\begin{table*}[t]
\caption{Pass@$k$ and Recall@$k$ of LLMs on \bench. The bold values indicate top-1 results.}
\label{tab:main_results}
\resizebox{\linewidth}{!}{
\begin{tabular}{lc|cccc|cccc}
\toprule
LLMs & Size & Pass@1 & Pass@3 & Pass@5 & Pass@10 & Recall@1 & Recall@3 & Recall@5 & Recall@10 \\
\midrule
\rowcolor[rgb]{ .741,  .843,  .933}
\multicolumn{10}{c}{Local File (Infilling)} \\
\midrule
gpt-4 & N/A & \textbf{53.04} & \textbf{56.05} & \textbf{58.16} & \textbf{60.65} & \textbf{71.38} & \textbf{71.87} & 72.90 & 74.12 \\
gpt-3.5 & N/A & 44.50 & 45.48 & 47.56 & 49.85 & 64.46 & 68.15 & 69.22 & 70.78 \\
DeepSeek Coder & 33B & 46.32 & 53.35 & 56.39 & 59.75 & 67.67 & 71.56 & \textbf{73.31} & \textbf{76.13} \\
DeepSeek Coder & 6.7B & 40.82 & 48.13 & 51.44 & 55.11 & 66.27 & 68.33 & 71.09 & 74.36 \\
\midrule
\rowcolor[rgb]{ .741,  .843,  .933}
\multicolumn{10}{c}{Local File (Completion)} \\
\midrule
gpt-4 & N/A & \textbf{47.44} & \textbf{52.41} & \textbf{54.48} & \textbf{56.98} & \textbf{65.06} & \textbf{69.30} & \textbf{70.25} & 71.32 \\
gpt-3.5 & N/A & 40.50 & 45.48 & 47.56 & 49.85 & 60.77 & 64.69 & 66.12 & 67.62 \\
DeepSeek Coder & 33B & 41.78 & 48.45 & 51.36 & 54.60 & 63.58 & 66.12 & 68.65 & 71.71 \\
DeepSeek Coder & 6.7B & 36.13 & 43.12 & 46.25 & 50.02 & 61.00 & 64.51 & 66.29 & 68.91 \\
StarCoder 2 & 15B & 37.78 & 44.45 & 47.40 & 50.80 & 60.81 & 64.78 & 67.23 & 69.90 \\
StarCoder 2 & 7B & 32.82 & 39.29 & 42.28 & 45.77 & 59.71 & 63.02 & 65.84 & 68.83 \\
CodeLLaMa & 13B & 41.94 & 48.83 & 51.92 & 55.66 & 63.33 & 67.68 & 70.16 & \textbf{72.62} \\
CodeLLaMa & 7B & 39.75 & 46.80 & 49.97 & 53.80 & 60.53 & 65.48 & 67.79 & 70.76 \\
\midrule
\rowcolor[rgb]{ .741,  .843,  .933}
\multicolumn{10}{c}{Without Context} \\ \midrule
gpt-4 & N/A & \textbf{17.40} & \textbf{20.19} & \textbf{21.24} & 22.55 & 16.85 & 18.53 & 19.56 & 20.98 \\
gpt-3.5 & N/A & 13.98 & 16.38 & 17.51 & 19.01 & 14.90 & 16.69 & 17.06 & 17.95 \\
DeepSeek Coder & 33B & 14.99 & 18.74 & 20.82 & \textbf{23.43} & \textbf{19.03} & \textbf{21.67} & 23.09 & 25.02 \\
DeepSeek Coder & 6.7B & 12.54 & 17.15 & 19.41 & 22.38 & 17.03 & 18.90 & 20.27 & 22.63 \\
StarCoder 2 & 15B & 11.05 & 16.02 & 18.25 & 21.12 & 15.48 & 17.89 & 19.94 & 22.43 \\
CodeLLaMa & 13B & 13.39 & 17.93 & 20.28 & 23.39 & 18.05 & 21.46 & \textbf{23.39} & \textbf{25.94} \\
CodeLLaMa & 7B & 12.70 & 17.44 & 19.93 & 22.91 & 16.36 & 19.02 & 20.98 & 23.93 \\
\bottomrule
\end{tabular}}
\end{table*}

%% file: chapter/Discussion.tex
\section{Discussion}
\label{sec:discussion}

\noindent \textbf{Results on different program types.} Figure \ref{fig:different_program} shows Pass@1 of gpt-4 on different program types (\ie standalone and non-standalone). We have two observations from the results. \ding{182} Contexts are crucial to generating non-standalone functions. For example, adding local files improves the Pass@1 on non-standalone functions from 11.04 to 53.58. \ding{183} Contexts also benefit standalone functions. This is attributed to the domain knowledge within contexts, aiding LLMs in understanding requirements. \ding{184} There is considerable room for improving LLMs on both programs. How to effectively retrieve relevant contexts is a key problem.

\begin{figure}[t]
\centering
\includegraphics[width=\linewidth]{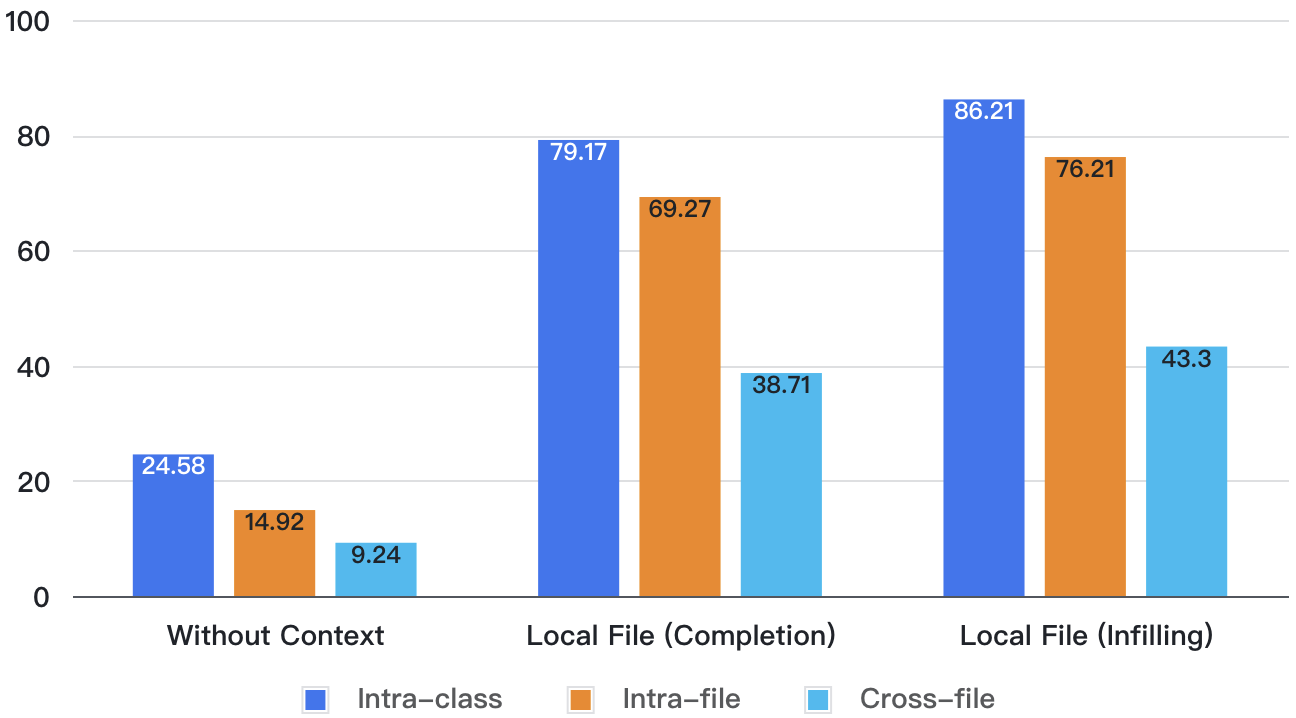}
\caption{Recall@1 of gpt-4 on different dependency types.}
\label{fig:different_depend}
\end{figure}

\noindent \textbf{Results on different dependency types.} Figure \ref{fig:different_depend} shows the Recall@1 of gpt-4 on different dependency types (\ie intra-class, intra-file, and cross-file). The results yield two insights. \ding{182} Without context, LLMs can reason about some simple dependencies from requirements (\eg initialization functions of returned objects), but still exhibit low Recall@1 values across three dependency types. \ding{183} With contexts, LLMs exhibit an improvement in generating dependencies. Nevertheless, LLMs have yet to grapple with generating dependencies, especially cross-file dependencies. As illustrated in Figure \ref{fig:error_case}, LLMs often ignore available dependencies defined in contexts.

\noindent \textbf{Data leakage.} 
Theoretically, all open-source code repositories may be included in the training data for LLMs. Consequently, there is a risk of data leakage where several repositories used to build \bench appear in the training data. We think this risk has only a slight impact on \bench due to three reasons. 
\ding{182} \bench contains new data, \ie manually written requirements. These requirements are never included in the training data.
\ding{183} Existing LLMs do not show overfitting tendencies to \bench. Based on the release dates of 8 LLMs (see Section \ref{sec:experiments:base_llms}), we divide \bench into 
two groups: unseen repositories released later than LLMs and potentially seen repositories released earlier than LLMs. The average difference of Pass@1 between the two groups is around 0.36. Compared to the average variations between LLMs (\eg 5.15 in Table \ref{tab:main_results}), 0.36 is slight. 
\ding{184} \bench is geared toward evaluating future LLMs. We release the links to our selected repositories and encourage practitioners to omit these repositories when collecting the training data for future LLMs.

\noindent \textbf{The bias of Recall@$k$.}
As stated in Section \ref{sec:DevEval:metric}, we develop a static analysis-based parser to extract dependencies in generated programs automatically. Because Python is a dynamically typed language, certain dependencies are only determined at runtime and may elude our parser. It may lead to lower Recall@$k$ than actual values.

To gauge the above bias, we randomly select 50 programs generated by gpt-4 and annotate dependencies with them by our parser and two human developers, respectively. Based on the human-annotated and auto-extracted dependencies, we compute two Recall@1 values. The bias of two Recall@1 values is 0.16. Compared to the average variations between LLMs (2.16 in Table \ref{tab:main_results}), 0.16 is slight. Consequently, we believe that Recall@$k$ can effectively rank different LLMs, notwithstanding its slight bias.

%% file: chapter/Related_Work.tex
\section{Related Work}
\label{sec:related_work}

\noindent \textbf{Large Language Models for Code Generation.}  
The rise of pre-training technology has brought new impetus to the field of code generation, both in academia and industry \cite{AlphaCode, Shen_CodeGen, CodeGen, InCoder}. In this context, more and more LLMs have emerged, achieving significant advancements in code generation, such as Codex \cite{Codex}, ChatGPT \cite{gpt-3.5}, CodeLlama \cite{CodeLLaMa}, DeepSeek Coder \cite{DeepSeek_Coder}, and StarCoder2 \cite{StarCoder-2}. 

To effectively steer LLMs in various code generation scenarios, some works focus on improving the prompt technologies by introducing specific patterns, \eg Structured Chain-of-Thought \cite{SCoT}, Self-planning \cite{Self-Planning}, Self-debug \cite{Self-Debug}, Self-collaboration \cite{Self-Collaboration}, and AceCoder \cite{AceCoder}. 

\noindent \textbf{Code Generation Benchmarks.} 
Early code generation benchmarks \cite{CoNaLA,Codex,MBPP,CERT} evaluate code generation on relatively Python functions, such as HumanEval \cite{Codex} and MBPP \cite{MBPP}. APPS \cite{APPS} evaluates code generation on more difficult competition-style problems. ClassEval \cite{ClassEval} evaluates LLMs on class-level code generation and contains 100 human-crafted self-contained Python classes. 
Concode \cite{Concode} and CoderEval \cite{CoderEval} further introduce non-standalone programs.

Compared to existing benchmarks, \bench aligns with real-world code repositories (\eg the distributions of code and dependency) and contains more comprehensive annotations (\eg reference dependencies).

We have also noticed that some benchmarks have recently been proposed for repository-level tasks. CrossCodeEval \cite{CrossCodeEval}, RepoBench \cite{RepoBench}, and RepoEval \cite{RepoCoder} are code completion benchmarks. They lack the necessary annotations (\eg natural language requirements) for code generation. SWE-bench \cite{SWE-bench} focuses on repairing repositories' issues by revising existing programs. In contrast, \bench is collected for code generation and aims to generate new programs based on requirements for a repository. \bench offers comprehensive annotations (\eg natural language requirements, original repositories, reference code, and reference dependencies).

%% file: chapter/Conclusion.tex
\section{Conclusion and Future Work}
\label{sec:conclusion}

This paper proposes a new code generation benchmark named \bench. Collected through a meticulous pipeline, \bench aligns with real-world code repositories in multiple dimensions, \eg real code distributions, sufficient dependencies, and real-scale repositories. We evaluate 8 popular LLMs in \bench. The results reveal the strengths and weaknesses of LLMs in real repositories. 
Compared to previous benchmarks, \bench offers a more challenging and practical evaluation scenario. We hope \bench can facilitate the applications of LLMs in practical repositories.

In the future, we will continue to update \bench, \eg multilingual testing samples, more projects, and more test cases. Besides, we will explore how to improve the performance of LLMs in context-based code generation, \eg retrieval-augmented and tool-augmented generation.

\section{Acknowledgments}
\label{sec:acknowledgment}

This research was supported by the National Natural Science Foundation of China (Nos. 62192731, 62152730).

\section{Limitations}
\label{sec:limitation}

This paper proposes a new code generation benchmark - \bench, which aligns with real-world code repositories. Based on \bench, we evaluate 8 popular LLMs and analyze their strengths and shortcomings. We think that \bench has three limitations. 

\ding{182} \bench is a monolingual benchmark (\ie requirements in English and code in Python) and ignores other languages. In practice, LLMs require understanding requirements in different natural languages (\eg Chinese, Spanish) and generating programs in various programming languages (\eg Java, C). Thus, we plan to build a multilingual \bench in future work. 

\ding{183}  As stated in Section \ref{sec:discussion}, Recall@$k$ values in \bench may have slight biases, \ie they may be slightly less than actual values. Because Python is a dynamically typed language, certain dependencies can only be identified at runtime and may elude our parser. 
To gauge the bias introduced by our parser, we manually annotate dependencies within 100 programs generated by gpt-4. Simultaneously, we employ the parser to extract dependencies in the same 50 programs. Based on the human-annotated and auto-extracted dependencies, we compute two Recall@1 values. The bias of two Recall@1 is 0.16. Compared to the average variations between LLMs (2.16 in Table \ref{tab:main_results}), 0.16 is slight. Consequently, the Recall@$k$ can effectively rank different LLMs, notwithstanding its slight bias.

\ding{184} In our experiments, we only consider the code contexts from local files. In the future, we will explore how to utilize broader contexts (\eg imported files, sibling files).

\section{Ethics Consideration}
\label{sec:ethics}

\bench is collected from real-world code repositories. We manually check all samples in \bench. We ensure all samples do not contain private information or offensive content. We ensure all programs in \bench are behaving normally and exclude any malicious programs.